\documentclass[lettersize,journal,siunitx]{IEEEtran}
\pdfoutput=1
\usepackage{amsmath,amsfonts}
\usepackage{array}
\usepackage{textcomp}
\usepackage{stfloats}
\usepackage{url}
\usepackage{verbatim}
\usepackage{graphicx}
\usepackage{cite}
\usepackage{subcaption}
\usepackage{xcolor}
\usepackage[pagebackref=true,breaklinks=true]{hyperref}
\hypersetup{
     colorlinks,
     linkcolor={red!75!black},
     citecolor={blue!75!black},
     urlcolor={blue!75!black},
}

\usepackage{xspace}
\usepackage[capitalise]{cleveref}
\usepackage{epstopdf}
\usepackage{amsmath,amssymb,booktabs,tabularx}
\usepackage{multirow}

\usepackage{algorithm}
\usepackage{algorithmicx}
\usepackage{algpseudocode}
\usepackage{bm}
\usepackage{microtype}
\usepackage{siunitx}
\usepackage{soul}

\newcommand{\ie}{i.\,e.,\xspace}
\newcommand{\eg}{e.\,g.,\xspace}

\begin{document}

\makeatletter
\let\old@ps@headings\ps@headings
\let\old@ps@IEEEtitlepagestyle\ps@IEEEtitlepagestyle
\def\confheader#1{%
	\def\ps@IEEEtitlepagestyle{%
		\old@ps@IEEEtitlepagestyle%
		\def\@oddhead{\strut\hfill#1\hfill\strut}%
		\def\@evenhead{\strut\hfill#1\hfill\strut}%
	}%
	\ps@headings%
}
\makeatother

\confheader{
	\parbox{20cm}{SUBMITTED TO IEEE TRANSACTIONS ON GEOSCIENCE AND REMOTE SENSING FOR POSSIBLE PUBLICATION.}
}

\title{AMD-HookNet for Glacier Front Segmentation}

\author{Fei Wu, Nora Gourmelon, Thorsten Seehaus, Jianlin Zhang, Matthias Braun, Andreas Maier, and Vincent Christlein
\thanks{Fei Wu and Jianlin Zhang are with the School of Electrical, Electronics and Communication Engineering, University of Chinese Academy of Sciences, 100049 Beijing, China. Also with the Key Laboratory of Optical Engineering at Institute of Optics and Electronics, Chinese Academy of Sciences, 610200 Chengdu, China (email: wufei171@mails.ucas.edu.cn)}
\thanks{Fei Wu, Nora Gourmelon, Andreas Maier and Vincent Christlein are with the Computer Science Department at Friedrich-Alexander-Universität Erlangen-Nürnberg, 91058 Erlangen, Germany.}
\thanks{Thorsten Seehaus and Matthias Braun are with the Geography and Geosciences Department at Friedrich-Alexander-Universität Erlangen-Nürnberg, 91058 Erlangen, Germany.}
\thanks{Corresponding authors: Jianlin Zhang; Vincent Christlein}
}

\maketitle

\begin{abstract}
Knowledge on changes in glacier calving front positions is important for assessing the status of glaciers. Remote sensing imagery provides the ideal database for monitoring calving front positions, however, it is not feasible to perform this task manually for all calving glaciers globally due to time-constraints. Deep learning-based methods have shown great potential for glacier calving front delineation from optical and radar satellite imagery. The calving front is represented as a single thin line between the ocean and the glacier, which makes the task vulnerable to inaccurate predictions. The limited availability of annotated glacier imagery leads to a lack of data diversity (not all possible combinations of different weather conditions, terminus shapes, sensors, etc.\ are present in the data), which exacerbates the difficulty of accurate segmentation. In this paper, we propose Attention-Multi-hooking-Deep-supervision HookNet (AMD-HookNet), a novel glacier calving front segmentation framework for synthetic aperture radar (SAR) images. The proposed method aims to enhance the feature representation capability through multiple information interactions between low-resolution and high-resolution inputs based on a two-branch U-Net. The attention mechanism, integrated into the two branch U-Net, aims to interact between the corresponding coarse and fine-grained feature maps. This allows the network to automatically adjust feature relationships, resulting in accurate pixel-classification predictions. Extensive experiments and comparisons on the challenging glacier segmentation benchmark dataset CaFFe show that our AMD-HookNet achieves a mean distance error of 438\,m to the ground truth outperforming the current state of the art by 42\,\%, which validates its effectiveness.
\end{abstract}


\section{Introduction}
\IEEEPARstart{G}{lacier} mass loss is one of the main contributors to global sea level rise~\cite{frederikse2020causes}. Many marine-terminating glaciers around the globe (Antarctica, the sub-Antarctic islands, Greenland, Russian and Canadian Arctic, Alaska, Patagonia) show considerable retreat and ice mass loss, with an observed acceleration in the last decade~\cite{vaughan2014observations,imbie2018mass,hugonnet2021accelerated}.
 Dynamic adjustments of glaciers, \eg acceleration of glacier flow and surface elevation changes, are closely related to frontal retreat~\cite{frank2022geometric}. Due to this relationship between frontal retreat and glacier dynamics, observation and accurate extraction of glacier calving front positions is essential for monitoring glacier dynamics~\cite{davari2020glacier}. Continuous monitoring of calving fronts enables a spatio-temporal quantification of frontal ablation, which is an important parameter of total glacier mass balance~\cite{liu2021automated}. To enable such a spatio-temporal quantification, it is desirable to extract glacier front positions on a regular basis adjusted to the expected front positions change -- from concurrent observation but also from archived satellite data. A large-scale manual glacier front extraction would result in unbearable time and cost expenses.

To cope with these challenges, many deep learning-based segmentation methods~\cite{baumhoer2019automated,zhang2019automatically,zhang2021automated,mohajerani2019detection,heidler2021hed,liu2021multiscale,periyasamy2022get,holzmann2021glacier,baumhoer2021environmental,dong2022automatic,marochov2021image,cheng2021calving,essd-14-4287-2022} have been investigated to extract glacier fronts automatically. However, most of these approaches~\cite{baumhoer2019automated,mohajerani2019detection, zhang2019automatically,liu2021multiscale,periyasamy2022get,holzmann2021glacier} have little or no additional interaction of information (\eg besides the skip connections) between the different layers/features in their inherited general U-Net architecture~\cite{ronneberger2015u}, which limits a closer fit of the predicted front to the manually delineated calving front. To this end, we propose a novel glacier segmentation method that integrates two individual U-Nets with inputs in different resolutions, where one of the inputs includes spatial surroundings of the other enabling a bigger field-of-view. The multi-scale inputs are integrated into a unified framework using deep supervision on the so called hook-up feature maps with attention mechanisms. Extensive experiments on the recently proposed challenging glacier segmentation benchmark dataset CaFFe\footnote{CaFFe Paper (DOI: \href{https://essd.copernicus.org/articles/14/4287/2022/}{10.5194/essd-14-4287-2022})}\footnote{CaFFe Dataset (DOI: \href{https://doi.org/10.1594/PANGAEA.940950}{10.1594/PANGAEA.940950})}~\cite{essd-14-4287-2022} demonstrate that our proposed method performs remarkably better than the baseline reducing the mean distance error of the front prediction from 753\,m to 438\,m. The CaFFe benchmark was chosen, as it comprises synthetic aperture radar (SAR) data (see Figure~\ref{fig2} for an example image and the corresponding landscape annotation). SAR data can help fill data gaps where optical data is insufficient (\eg polar nights), as seasonal changes can be tracked. However, due to speckle noise, the calving front is harder to delineate in SAR data than in optical data~\cite{baumhoer2018remote,tedesco2015remote}. Therefore, better deep learning models (like the proposed AMD-HookNet) are needed. To verify the necessity of each component of our proposed enhancements to the original HookNet~\cite{van2021hooknet}, several ablation studies are undertaken. In particular, the proposed different network components are introduced to the HookNet~\cite{van2021hooknet} one after another, and the resulting architectures are trained and evaluated.

The remainder of the paper is structured as follows: we briefly review the related works on glacier segmentation in \cref{sec:related_work}. The detailed network architecture and components of the proposed method are demonstrated in \cref{sec:methodology}. \cref{sec:evaluation} presents the details of the experimental setup of the proposed method and gives the evaluation results and analysis on the CaFFe dataset. In addition, we illustrate several ablation studies on the proposed components. \cref{sec:conclusion} concludes the paper.

\section{Related Work}\label{sec:related_work}
Most of the presented glacier segmentation methods are based upon follow-up works of the classical U-Net architecture~\cite{ronneberger2015u}, which consists of an encoder aiming to capture context and a symmetric reverse decoder aiming to achieve accurate pixel-classification. For example, Baumhoer et al.~\cite{baumhoer2019automated} modify the U-Net architecture to have less parameters to segment into land-ice and ocean. In contrast, Mohajerani et al.~\cite{mohajerani2019detection} explore the optimal U-Net architecture for the case of lower resolution multispectral Landsat images and Liu et al.~\cite{liu2021multiscale} incorporate dilated convolutions and residual connections based on the U-Net architecture to automatically map glacier contours. Moreover, Zhang et al.~\cite{zhang2019automatically} use larger convolutional kernel sizes in U-Net to smoothly depict the glacier fronts of Jakobshavn Isbræ while Periyasamy et al.~\cite{periyasamy2022get} make an effort to fully optimize the U-Net architecture for glacier segmentation. A different approach is taken by Heidler et al.~\cite{heidler2021hed} who propose a multitask learning paradigm for information interacting between glacier fronts and glacier masks in U-Net.

Different from the U-Net applications listed above, Baumhoer et al.~\cite{baumhoer2021environmental} analyze the circum-Antarctic calving front changes over the last two decades while Dong et al.~\cite{dong2022automatic} develop a binary classification algorithm to distinguish between ice mélange and grounded ice based on game theory in high-resolution Digital Elevation Model (DEM) data. Instead of directly segmenting the entire image into the desired classes, Marochov et al.~\cite{marochov2021image} propose a two-phase deep learning workflow based on the VGG architecture~\cite{simonyan2014very} to efficiently classify every single pixel in each  glacial landscape separately. Some works explore the DeepLabv3~\cite{chen2017rethinking} architecture for glacier segmentation. While Cheng et al.~\cite{cheng2021calving} modify the original DeepLabV3+Xception~\cite{chen2018encoder} with improved image preprocessing and postprocessing to analyze the seasonal and regional trends in tidewater glaciers of Greenland, Zhang et al.~\cite{zhang2021automated} combine the DeepLabV3 architecture with various popular backbones~\cite{he2016deep,yu2017dilated,howard2017mobilenets} to fully evaluate its potential.

In the field of medical image segmentation, Rijthoven et al.~\cite{van2021hooknet} proposed the HookNet for the segmentation of tissue in histopathology images.
The basic idea is to use two inputs, one detailed view that is going to be segmented, and another view of the same input but with a larger context and less details that helps to get the `big picture' around the detailed view. 
This idea is realized in a segmentation framework consisting of two branches. Each branch inherits the classical U-Net architecture~\cite{ronneberger2015u}. The aligning of the context branch with the target branch is referred to \emph{hooking}. Similar to tissue segmentation, we face the problem of small glacier fronts where additional context may help to improve the fine-grained segmentation. The success of this method inspired our own exploration and further improvements for its use in glacier front segmentation.

A key improvement is the incorporation of attention to the hooking mechanism that increases the interaction between both branches. Attention refers to a mechanism that aims to focus on the critical content of the input, which is of great importance for information extraction and data processing, while suppressing redundant signals~\cite{woo2018cbam}. It adjusts the relationship among different input components with learnable weights. It has significant applications in natural language processing~\cite{hu2019introductory,galassi2020attention,liu2021attention,shen2018disan}, image classification~\cite{touvron2021training,liu2021swin,lu2021soft}, semantic segmentation~\cite{wang2021crossformer,lee2022mpvit,ali2021xcit,wang2021pyramid,liu2021swin}, and recently flourished Transformer~\cite{vaswani2017attention,liu2021swin,touvron2021training} architectures. For glacier-related works, Holzmann et al.~\cite{holzmann2021glacier} propose attention gates to learn key information in the skip connections of the U-Net~\cite{ronneberger2015u}, and use a distance-weighted loss function to deal with the pixels' class imbalance in favor of glacier front segmentation. Chu et al.~\cite{chu2022glacier} use attention in combination with the segmentation network DeepLabV3+~\cite{chen2018encoder} while Yan et al.~\cite{yan2021glacier} propose spectral and spatial attention modules in the U-Net~\cite{ronneberger2015u} for glacier classification.

\begin{figure*}[t]
\centering
\includegraphics[width=\textwidth]{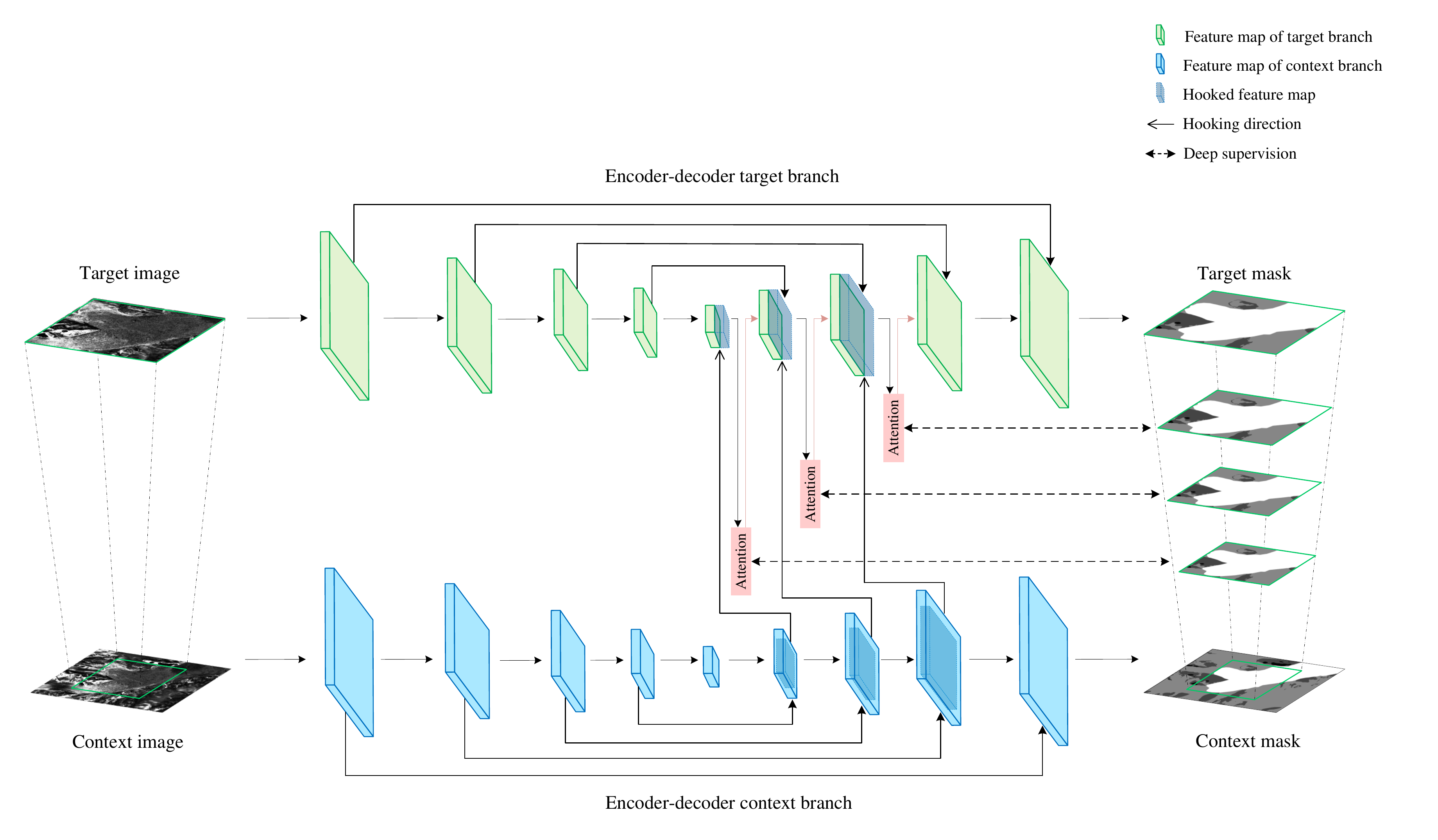}
\caption{Architectural details of the proposed network. It consists of a target branch and a context branch, where the target input is center-cropped from the context input but equipped with a higher image resolution, the context input has more surroundings and is downsampled to match the size of the target input leading to a lower resolution. The two branches are linked by three attention-hooking modules, supervised and optimized by correspondingly downsampled ground truth.}\label{fig1}
\end{figure*}

\section{Methodology}\label{sec:methodology}
The U-Net~\cite{ronneberger2015u} architecture is a popular segmentation approach consisting of an encoder and a reverse decoder. There is a series of follow-up U-Nets for glacier segmentation~\cite{baumhoer2019automated,heidler2021hed,holzmann2021glacier,mohajerani2019detection,liu2021multiscale,zhang2019automatically,periyasamy2022get,essd-14-4287-2022}. Most of them~\cite{baumhoer2019automated,mohajerani2019detection,heidler2021hed,holzmann2021glacier,zhang2019automatically,periyasamy2022get,cheng2021calving,essd-14-4287-2022} implement binary segmentation while others~\cite{marochov2021image,essd-14-4287-2022} implement multi-class segmentation. Direct implementation of edge detection is prone to cause inaccurate and blurry predictions~\cite{heidler2021hed} due to the imbalanced class distribution in limited data amount~\cite{fidon2017generalised}. As a result, most works~\cite{baumhoer2019automated,periyasamy2022get,holzmann2021glacier,mohajerani2019detection,liu2021multiscale,zhang2019automatically} begin with zone segmentation followed by postprocessing to delineate glacier calving fronts. We follow this methodology in this paper as the multi-class baseline~\cite{essd-14-4287-2022} showed a lower MDE over the complete test set compared to the binary baseline. The success of the HookNet~\cite{van2021hooknet} showed that the information exchange between the fine-grained feature maps and the coarse-grained feature maps has a large impact on the segmentation performance, which inspired us to explore an extended hooking mechanism for glacier segmentation. In this work, we propose a unified glacier segmentation network that integrates \textbf{a}ttention mechanisms into \textbf{m}ulti-hooking U-Nets with \textbf{d}eep supervision on the feature pyramid, dubbed as AMD-HookNet. By introducing a multiple attention-hooking mechanism based on a hierarchical attention scheme, the pyramids of fine-grained visual information from high-resolution patches and contextual-coarse visual information from low-resolution patches are hooked to produce the final output. In combination with deep supervision, applied to the downsampled ground truth, the learned attention effectively accelerates information interaction between fine-grained and coarse-grained features contributing to a promising performance. In this way, our proposed method takes full advantage of the information interaction from the coarse-grained image patch which is equipped with more context around the high-resolution image patch to the fine-grained image patch. The effectiveness of the proposed method is verified by extensive experimental results on the recently proposed challenging glacier segmentation benchmark dataset CaFFe~\cite{essd-14-4287-2022}.

The key components of our proposed AMD-HookNet are: (1)~the network architecture with the attention module and the multi-hooking mechanism, (2)~the deep supervision on attention-hooking feature pyramid, and (3)~the joint loss for the network optimization.

\subsection{Network Architecture}
Information interacting between coarse and fine-grained feature maps improves the network's ability to recognize and classify corresponding objects, as shown in various computer vision tasks~\cite{van2021hooknet,li2019siamrpn++,wosner2021object,zhao2015dirichlet}. 
We develop a cross-resolution segmentation method based on two connected U-Nets. The overview of our proposed method that integrates the attention mechanism into multi-hooking U-Nets with deep supervision on the feature pyramid (AMD-HookNet) is shown in \cref{fig1}. Similar to the original HookNet~\cite{van2021hooknet}, it consists of a context branch and a target branch. Each branch inherits the typical encoder-decoder U-Net architecture. The input of the target branch is center-cropped from the context branch but has fine-grained high-resolution information, while the input of the context branch is downsampled to match the size of the target input leading to low-resolution contextual information. Each basic convolutional block consists of two groups of a convolutional layer with a kernel size of 3 and padding size of 1, batch normalization, ReLU, and a max-pooling/de-convolution layer for the encoder/decoder, respectively. The basic number of channels for each convolution filter is 32 and increases in multiples of the network depth until the bottleneck is reached. \cref{tab1} lists the channel specifications for the two branches of AMD-HookNet, where feature size refers to the size of the activation maps (volume). The feature resolution ratio from target branch to context branch within convolutional blocks follows the alignment paradigm as:
\begin{equation}
     \mathrm{r^{B}_{t}}=\mathrm{2\times{r^{B}_{c}}}
\end{equation}
where $\mathrm{r_{t}}$ and $\mathrm{r_{c}}$ denote the resolution of the same region in target branch ($\mathrm{t}$) and context branch ($\mathrm{c}$) correspondingly, and $\mathrm{B\in{\{1,2,...,8,9\}}}$ denotes the depth of the convolutional blocks in each branch. 

\begin{table*}[t]
	\centering
	\caption{Layer specifications of AMD-HookNet.}
 \begin{tabular*}{\textwidth}{@{\extracolsep{\fill}}clccccc@{\extracolsep{\fill}}}
	\toprule
        &\multirow{2}*{}&\multicolumn{2}{c}{Feature size} \\
        &&Context branch&Target branch \\
        \midrule
	&Input&$288\times288\times1$&$288\times288\times1$ \\
        \midrule
        \multirow{5}{*}{Encoder}&Convolution block 1&$288\times288\times32$&$288\times288\times32$ \\
        &Pooling&$144\times144\times64$&$144\times144\times64$ \\
        &Convolution block 2 + Pooling&$72\times72\times128$&$72\times72\times128$ \\
        &Convolution block 3 + Pooling&$36\times36\times256$&$36\times36\times256$ \\
        &Convolution block 4 + Pooling&$18\times18\times320$&$18\times18\times320$ \\
        \midrule
        \multirow{5}{*}{Decoder}&Convolution block 5 + Upsampling&$36\times36\times256$&$36\times36\times256$ \\
        &Convolution block 6 + Upsampling&$72\times72\times128$&$72\times72\times128$ \\
        &Convolution block 7 + Upsampling&$144\times144\times64$&$144\times144\times64$ \\
        &Convolution block 8 + Upsampling&$288\times288\times32$&$288\times288\times32$ \\
        &Convolution block 9&$288\times288\times4$&$288\times288\times4$ \\
        \bottomrule
 \end{tabular*}
 \label{tab1}
\end{table*}

\subsubsection{Attention Module}
\label{subsubsec:attention_module}
The attention mechanism is the fundamental component in the design of the AMD-HookNet. Similar to the previous works~\cite{vaswani2017attention,dosovitskiy2020image,lin2021swintrack}, given queries $Q$, keys $K$ and values $V$, the attention function is the following scaled dot-product attention:
\begin{multline}
    \mathrm{Attention}(Q(\bm{M}),K(\bm{M}),V(\bm{M}))=\\ \mathrm{SoftMax}(\frac{Q(\bm{M})\times{K(\bm{M})^\top}}{\sqrt{d_{k}}})V(\bm{M})
\label{equ2}
\end{multline}
where $Q$, $K$ and $V$ are all derived from the same input matrix $\bm{M}$, $d_{k}$ denotes the dimension of $K$ and is used to scale the function. Self-attention is implemented on all hooked features, with the aim of guiding and focusing on valuable information during the cross-interaction between fine-grained target and coarse-grained context. This avoids simple feature fusion using fixed weights, which facilitates to improve the segmentation performance.

\subsubsection{Multi-Hooking Mechanism}
We propose to combine coarse-contextual information from the context branch into the target branch via attention-hooking operations introduced from corresponding convolutional blocks. Concretely, multiple attention-hooking operations are proposed to increase the efficiency of information exchange. It can be summarized as:
\begin{multline}
    \mathrm{Attention\mbox{-}hooking}^{D}=\\ \mathrm{Attention}(\mathrm{Concat}(\mathrm{target}^{D-1}_{r_{t}}, \mathrm{context}^{D}_{r_{c}}))
\label{equ3}
\end{multline}
where $r_{t}$ and $r_{c}$ denote the resolution of the same region in target branch ($t$) and context branch (${c}$) correspondingly. To achieve feature alignment, the center-cropped feature maps, $\mathrm{context}^{D}_{r_{c}}$, of the context branch are hooked and concatenated to the feature maps, $\mathrm{target}^{D-1}_{r_{t}}$, of the target branch. The hooking features have the same spatial size in the identical depth $D$ of the upsample convolutional blocks in the decoder, where $D\in{\{1,2,3\}}\subseteq{B}$. The attention-hooking operation computes attention maps using $Q$, $K$, and $V$ which are derived from the hooking feature: $\mathrm{Concat}(\mathrm{target}^{D-1}_{r_{t}}, \mathrm{context}^{D}_{r_{c}})$, which is regarded as $\bm{M}$ in~\cref{equ2}. Attention-hooking allows the network to focus on interleaving the contextual information and the fine-grained information from the context branch and the target branch, respectively, aiming to learn to enhance the feature representation capability.

Note that we modified the original HookNet architecture~\cite{van2021hooknet} to be specific to our task. First, the resolution ratio from the target input to the context input is changed from 4:1 to 2:1 to enhance the calculation efficiency, where 4:1 is used in the original HookNet~\cite{van2021hooknet}. For example, the given size of the target image in this work is $288\times288$, under the original HookNet settings, the size of the context image would be $1152\times1152$. For a complete image with an average image size of about $2016\times2016$ pixels in the CaFFe training set, the original HookNet~\cite{van2021hooknet} setting would result in zero-padding three-quarters of the context image patches when the patch is taken from the borders of the original image, which introduces interference and influences the class distribution of the data. In order to realize feature alignment, the hooked position is converted from the output of the second upsample layer to the first upsample layer. The additional multi-hooking positions in the proposed method are built upon the feature maps of the first and the second upsample layers in the target branch, and the corresponding aligned features derive from the second and the third upsample layer in the context branch. Second, we use convolutional layers with a kernel size of 3 \emph{with} padding in our proposed architecture, unlike the original HookNet~\cite{van2021hooknet}, which uses convolution without padding. This significantly reduces the amount of training data needed because for the original HookNet~\cite{van2021hooknet} the output dimensions of a patch would only be $74\times74$ covering a significantly less area of the $288\times288$ large patch. Therefore, for the entire CaFFe dataset~\cite{essd-14-4287-2022}, we need about 19 times fewer image patches if we use convolutional padding.

\subsection{Deep Supervision}
Lee et al.~\cite{lee2015deeply} proposed deep supervision to enhance the network learning and generalization capabilities. The outputs generated from the intermediate layers are supervised by the ground truth to monitor the training process and give feedback to the earlier layers for further model updating. In this work, deep supervision on the attention-hooking feature pyramid is used to help the proposed method to increase learning effectiveness and extend generalization capabilities. In particular, the ground truth is downsampled to align with the outputs of the attention-hooking feature pyramid infusing additional information and knowledge to the earlier convolutional blocks. The difference between downsampled ground truth images and corresponding attention-hooking predictions are regarded as additional loss terms to form the final joint loss. This enables the network to smoothly adapt valuable attention features of different receptive fields in variable upsample convolutional structures.

\subsection{Loss Function}
We apply deep supervision to the attention-hooking feature pyramid to fully optimize the entire network. For each encoder-decoder branch, the joint loss which consists of the cross-entropy loss and the dice loss with balanced coefficients is used for model weights optimization. The final loss is defined as follows:
\begin{equation}
\mathrm{Loss}=\lambda{_1}\mathrm{Loss_{t}}+\lambda{_2}\mathrm{Loss_{c}}+\lambda{_3}\mathrm{Loss_{deep}}
\label{eq7}
\end{equation}
\begin{equation}
    \mathrm{Loss_{t}}=\mathrm{CE}(o_\mathrm{t},y_\mathrm{t})+\mathrm{Dice}(o_\mathrm{t},y_\mathrm{t})
\end{equation}
\begin{equation}
    \mathrm{Loss_{c}}=\mathrm{CE}(o_\mathrm{c},y_{c})+\mathrm{Dice}(o_\mathrm{c},y_\mathrm{c})
\end{equation}
\begin{multline}
    \mathrm{Loss_{deep}}=\sum_{D=1}^{3}\bigl(\mathrm{CE}(\mathrm{Attention\mbox{-}hooking}{^{D}_{up}},y_\mathrm{t}^{D})+ \\ \mathrm{Dice}(\mathrm{Attention\mbox{-}hooking}{^{D}_{up}},y_\mathrm{t}^{D})\bigr)
\end{multline}
where $o_{*}$ and $y_{*}$ denote the final classification layer and the corresponding ground truth of the target branch ($\mathrm{t}$) and context branch ($\mathrm{c}$), respectively. $\mathrm{Loss_{deep}}$ denotes the deep supervision on the upsampled ($*_{up}$) feature maps of  $\mathrm{Attention\mbox{-}hooking}^{D}$ and corresponding target ground truth $y_\mathrm{t}^{D}$ at upsample depth $D\in{\{1,2,3\}}\subseteq{B}$. $\mathrm{Loss_t}$ denotes the target branch loss. $\mathrm{Loss_{c}}$ denotes the context branch loss. $\lambda{_1},\lambda{_2},\lambda{_3}$ are all hyper-parameters.

\begin{figure*}
\centering
\subcaptionbox{SAR images of Columbia (left) and Mapple (right) glaciers (details below) \label{fig:sar}}{\includegraphics[width=0.60\textwidth]{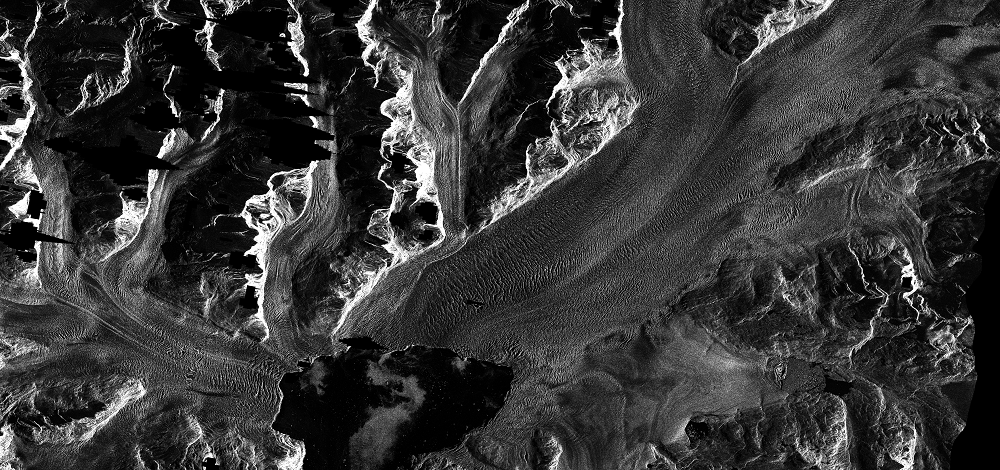} \hspace{10pt}\includegraphics[width=0.266\textwidth]{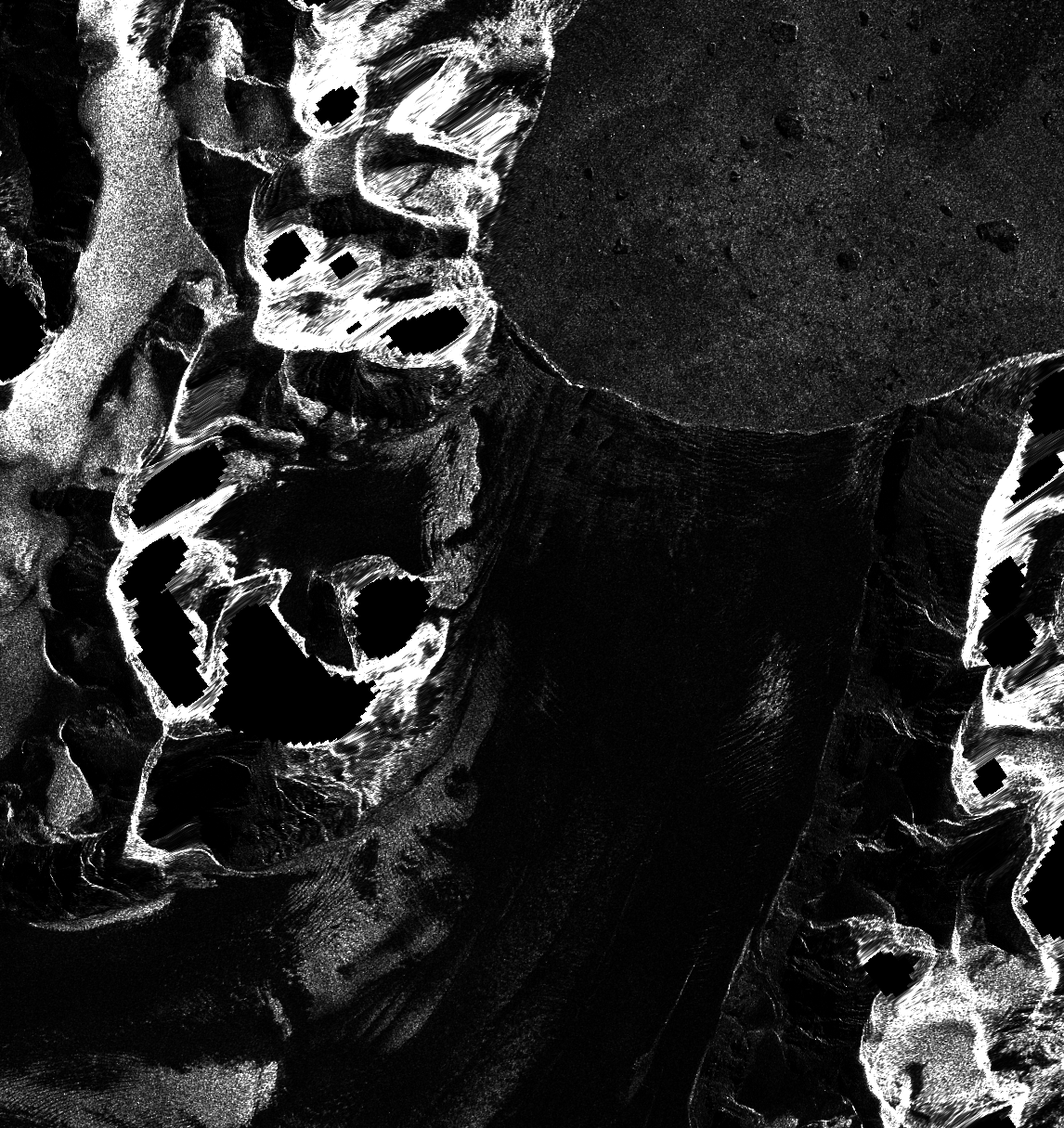}} \\
\subcaptionbox{Ground truth labels\label{fig:gt}}{\includegraphics[width=0.60\textwidth]{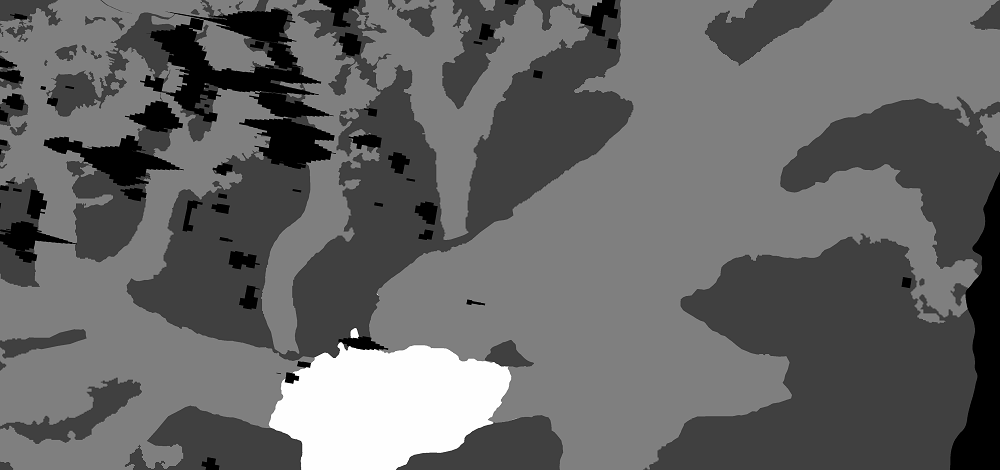} \hspace{10pt}\includegraphics[width=0.266\textwidth]{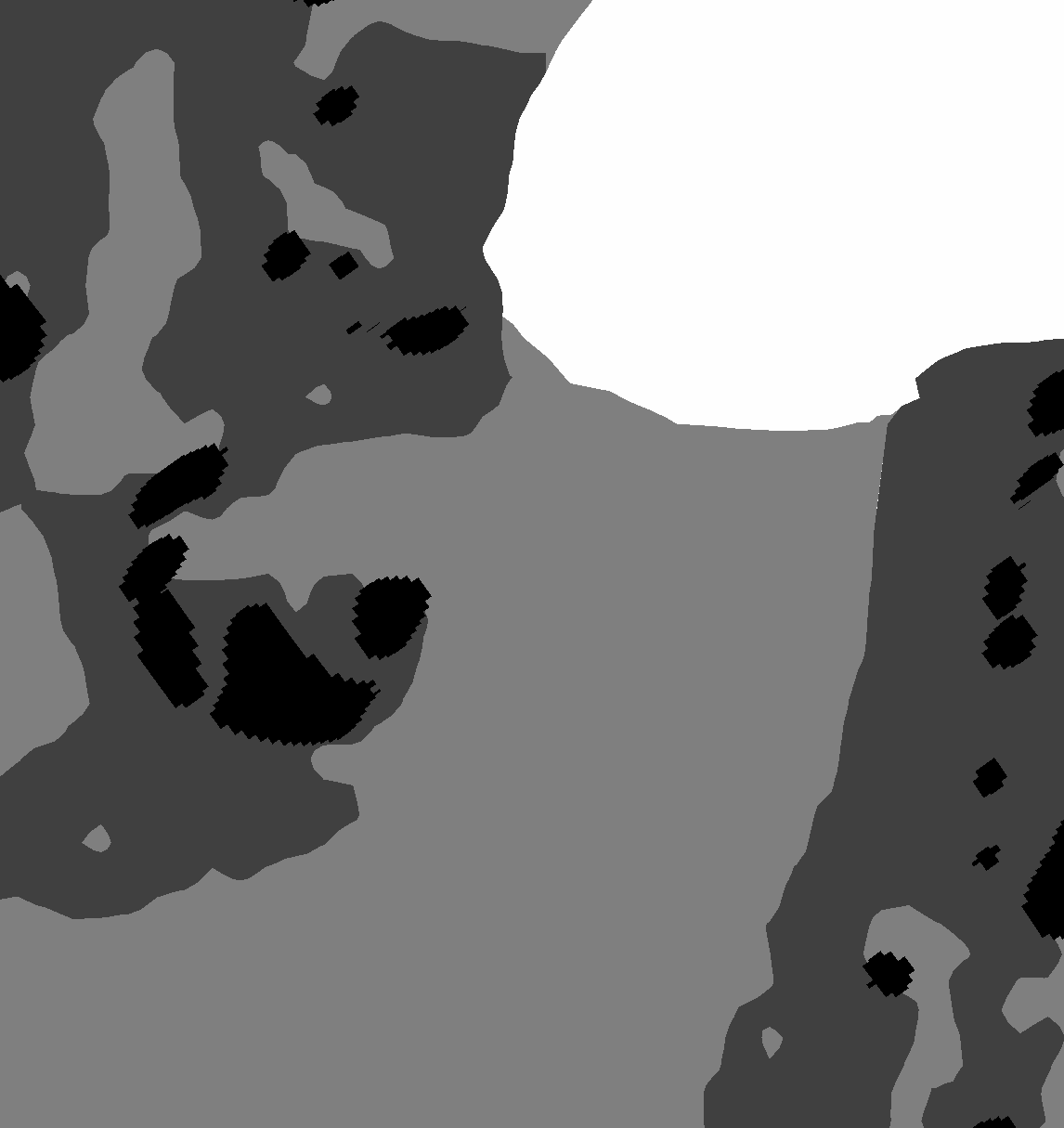}} \\
\subcaptionbox{HookNet output\label{fig:hooknet}}{\includegraphics[width=0.60\textwidth]{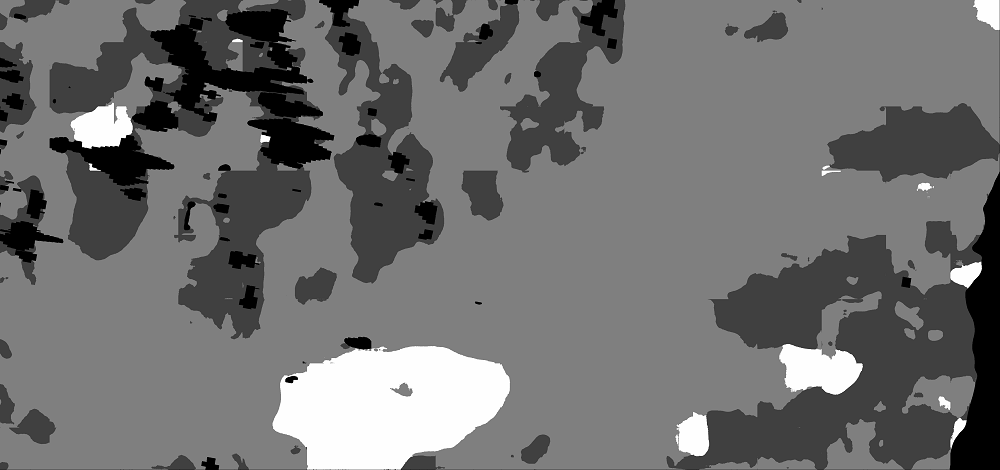} \hspace{10pt}
\includegraphics[width=0.266\textwidth]{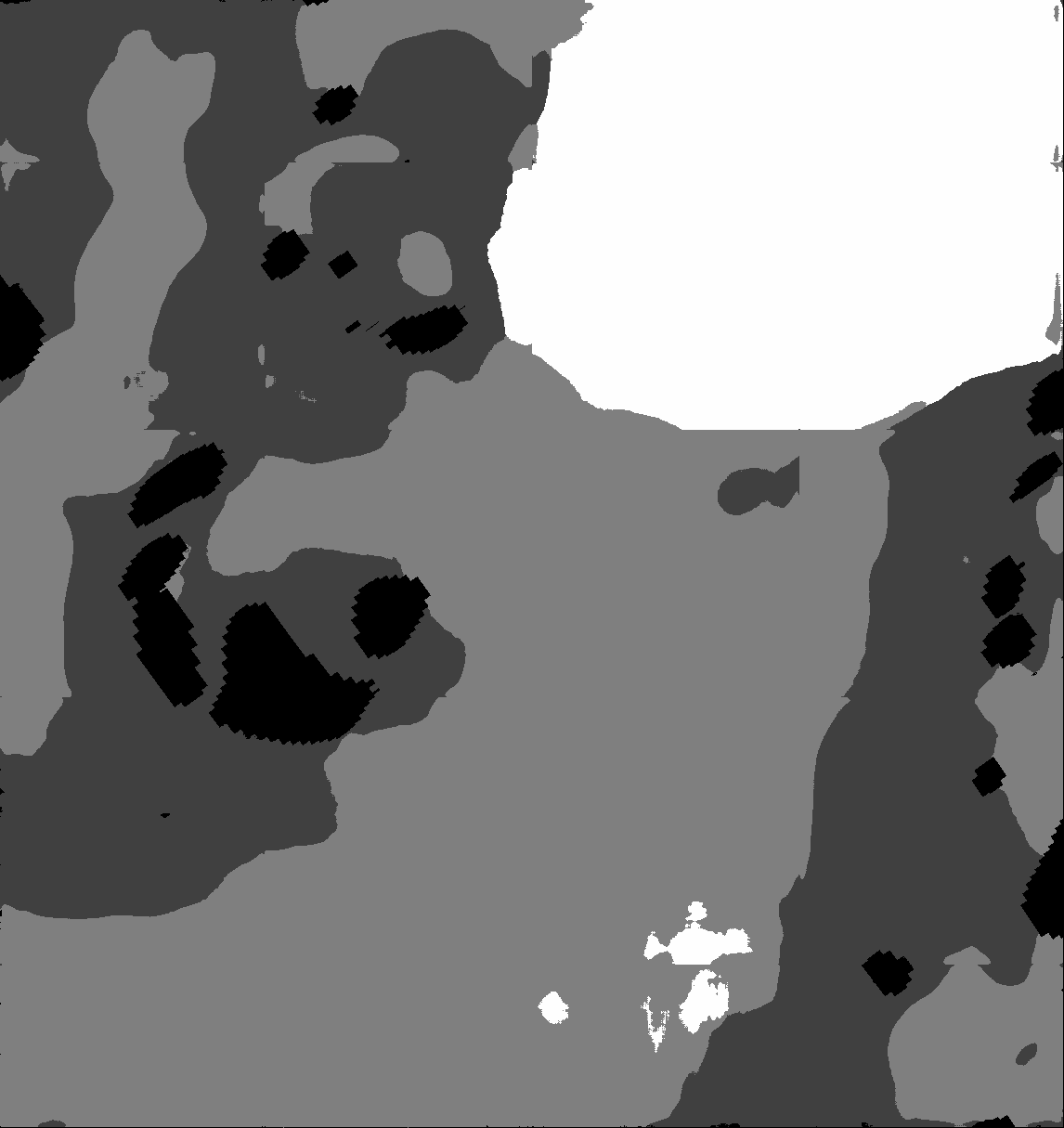}} \\
\subcaptionbox{AMD-HookNet output\label{fig:amdhook}}{\includegraphics[width=0.60\textwidth]{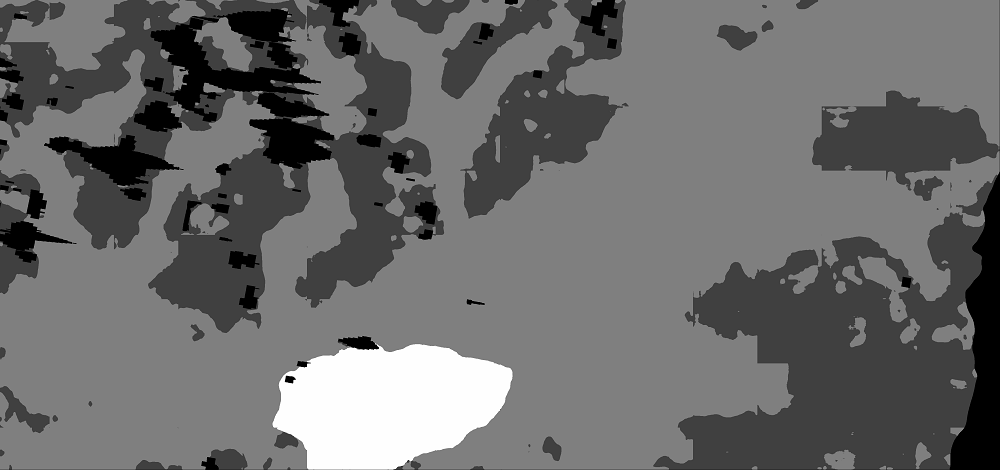} \hspace{10pt}
\includegraphics[width=0.266\textwidth]{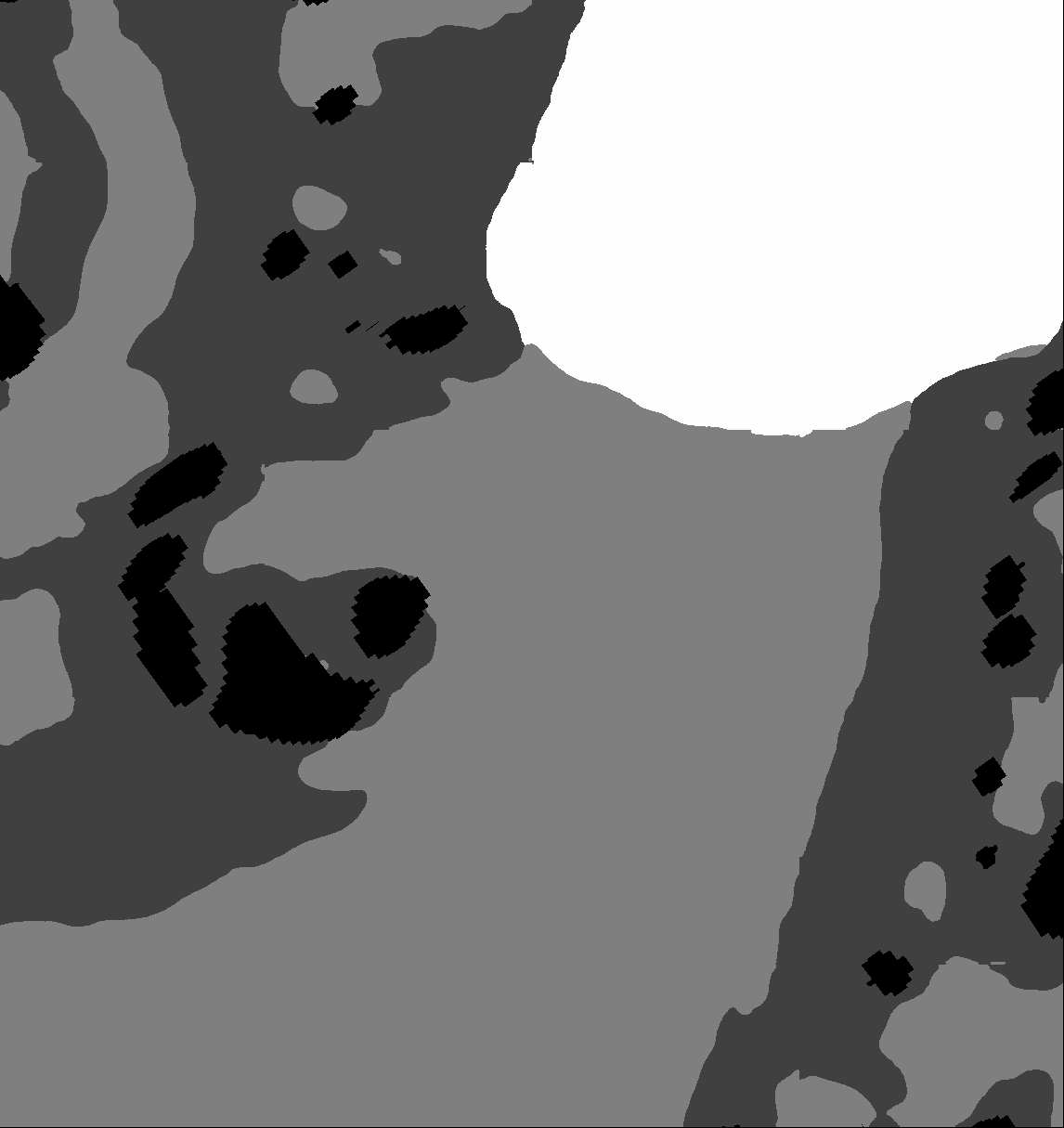}} \\
\caption{Qualitative comparison of the segmentation maps of \subref{fig:gt} ground truth, \subref{fig:hooknet} HookNet, and \subref{fig:amdhook} AMD-HookNet applied on the \subref{fig:sar} SAR images of the (left) Columbia Glacier (12th of Augest 2011, TanDEM-X) and (right) Mapple Glacier (25th of Decemeber 2010, the TerraSAR-X). For the output maps, white represents the ocean, light gray the ice, dark gray the rock outcrop and black a ``no data available'' region.}
\label{fig2}
\end{figure*}
\begin{figure*}
\centering
\subcaptionbox{\label{well1}}{\includegraphics[width=0.60\textwidth]{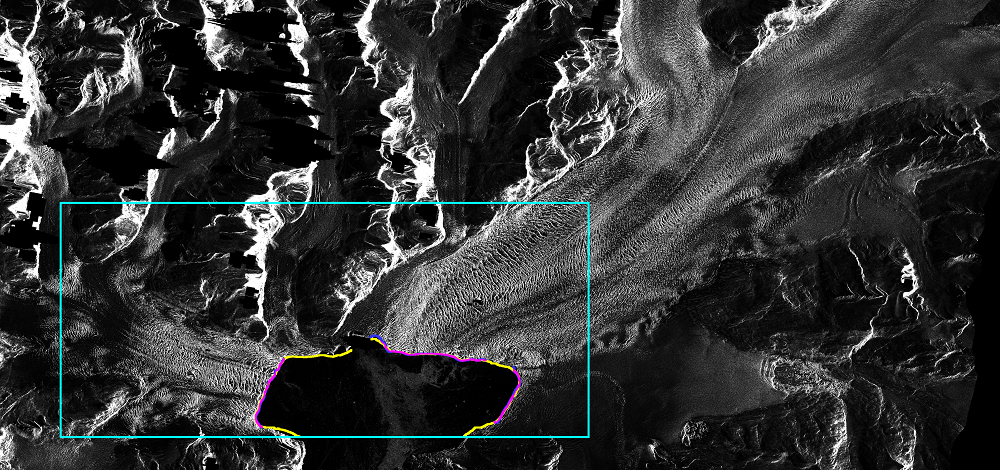}} \hspace{10pt}
\subcaptionbox{\label{well2}}{\includegraphics[width=0.266\textwidth]{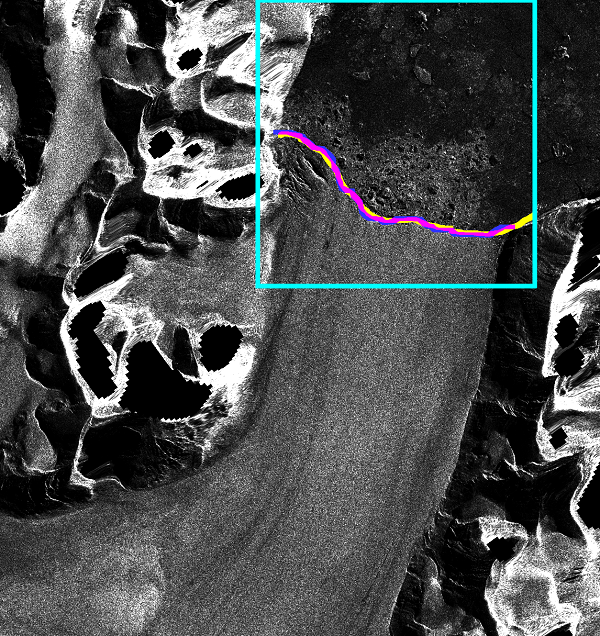}} \\
\subcaptionbox{\label{fail1}}{\includegraphics[width=0.60\textwidth]{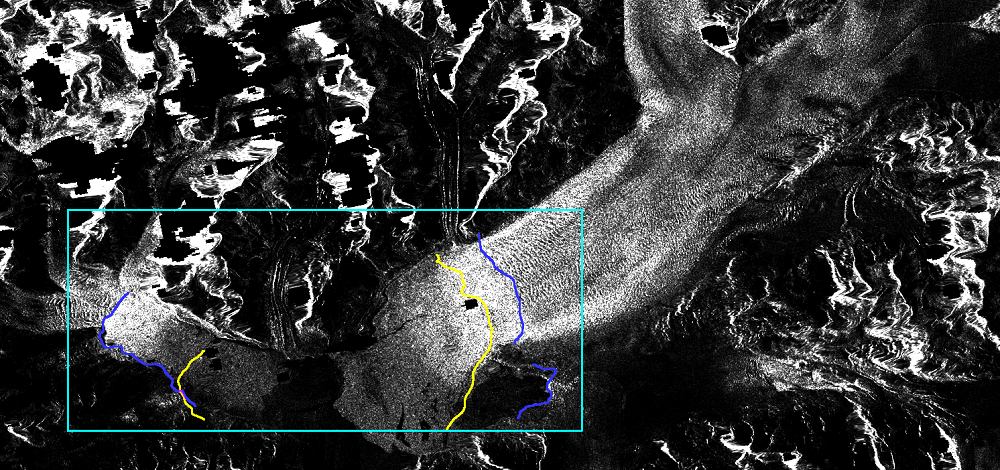}} \hspace{10pt}
\subcaptionbox{\label{fail2}}{\includegraphics[width=0.266\textwidth]{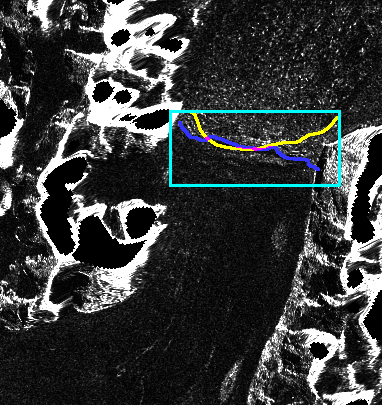}}
\caption{Visualization of the glacier front detection of AMD-HookNet on the test set. The blue, yellow, and pink colors represent the ground truth, the prediction, and the overlap between the ground truth and prediction, respectively. The turquoise rectangle is the bounding box specified separately for each glacier to obtain only the calving front of the observed dynamic glacier. \subref{well1} is an image of the Columbia Glacier acquired on the 19th of March 2012 by the TanDEM-X (TDX) satellite. \subref{well2} is an image of the Mapple Glacier acquired on the 2nd of November 2019 by the TerraSAR-X (TSX) satellite. \subref{fail1} is an image of the Columbia Glacier acquired on the 6th of January 2018 by the Sentinel-1 satellite. \subref{fail2} is an image of the Mapple Glacier acquired on the 8th of January 2020 by the Sentinel-1 satellite.}
\label{fig3}
\end{figure*}
\begin{table*}[t]
	\centering
	\caption{Comparisons between baseline and AMD-HookNet based on the evaluation metrics precision, recall, F1-score, IoU.}
	\begin{tabular*}{\textwidth}{@{\extracolsep{\fill}}lccccccccc@{\extracolsep{\fill}}}
	\toprule
    Scope&Method&{\bfseries Precision$\uparrow$}&{\bfseries Recall$\uparrow$}&{\bfseries F1-score$\uparrow$}&{\bfseries IoU$\uparrow$}\\
	\midrule
	\multirow{2}*{All}&Baseline&84.2$\pm$0.5&79.6$\pm$0.9&80.1$\pm$0.5&69.7$\pm$0.6 \\
	&Ours&\bfseries{85.0$\pm$0.6}&\bfseries{85.0$\pm$0.7}&\bfseries{84.3$\pm$0.7}&\bfseries{74.4$\pm$1.0} \\
	\multirow{2}*{NA Area}&Baseline&\bfseries{99.5$\pm$0.1}&91.2$\pm$1.3&\bfseries{94.8$\pm$0.8}&\bfseries{90.9$\pm$1.3} \\
	&Ours&93.9$\pm$0.4&\bfseries{94.4$\pm$0.9}&94.1$\pm$0.4&89.1$\pm$0.6 \\
	\multirow{2}*{Rock Outcrop}&Baseline&\bfseries{82.0$\pm$0.5}&59.6$\pm$1.3&67.9$\pm$0.8&53.5$\pm$0.7 \\
	&Ours&78.7$\pm$1.0&\bfseries{71.0$\pm$3.7}&\bfseries{73.9$\pm$2.3}&\bfseries{59.8$\pm$2.7} \\
	\multirow{2}*{Glacier}&Baseline&74.5$\pm$0.7&\bfseries{89.5$\pm$1.1}&80.9$\pm$0.3&68.5$\pm$0.5 \\
	&Ours&\bfseries{81.1$\pm$1.7}&85.2$\pm$1.8&\bfseries{82.0$\pm$1.1}&\bfseries{70.1$\pm$1.6} \\
	\multirow{2}*{Ocean and Ice Melange}&Baseline&80.9$\pm$2.2&78.3$\pm$3.1&76.8$\pm$1.6&66.0$\pm$1.5 \\
	&Ours&\bfseries{85.8$\pm$3.1}&\bfseries{90.0$\pm$0.9}&\bfseries{86.7$\pm$1.6}&\bfseries{78.2$\pm$2.3} \\
	\bottomrule
	\end{tabular*} \label{tab2}
\end{table*}

\section{Evaluation}\label{sec:evaluation}
In this section, we give the evaluation results and analysis to validate the effectiveness of the suggested improvements. Several competing approaches, as well as the proposed method, are evaluated on the challenging glacier segmentation benchmark dataset CaFFe~\cite{essd-14-4287-2022}. Performance differences of machine learning models originate from two sources: the data (training and test set) and the model architecture. As the focus of this research lies on advancing deep learning architectures for calving front delineation, we fix one of these sources by training and evaluating all competing model architectures on the same data -- a so-called benchmark dataset. In this way, we can guarantee that performance differences originate from the model architectures. For a fair comparison, we use the same visualization and experimental evaluation criteria as in~\cite{essd-14-4287-2022} to summarize and explicate the experimental details.

\subsection{Dataset}
Our experiments are implemented on CaFFe~\cite{essd-14-4287-2022} which is a benchmark dataset of SAR images from Antarctica, Greenland and Alaska with fully manual predefined labels to support scientific research in detecting glacier calving fronts. The SAR imagery of CaFFe~\cite{essd-14-4287-2022} stems from six different SAR satellite sensors (ERS-1/2, Envisat, RADARSAT-1, ALOS Phased Array L-band Synthetic Aperture Radar (ALOS PALSAR), TerraSAR-X (TSX) and TanDEM-X (TDX), and Sentinel-1A/B) covering the period from 1995 to 2020 with different spatial resolutions, frequencies, and signal-to-noise characteristics.

CaFFe~\cite{essd-14-4287-2022} already provides a split into a training and a glacier-independent test set with different class distributions, which makes this dataset challenging. We use this split accordingly to allow comparability with the given baseline. There are two sets of labels and corresponding two baseline models in~\cite{essd-14-4287-2022}: One binary segmentation model of “front” labels and one multi-class segmentation model of the “zones” labels which consist of four categories: ocean and ice-melange, rock outcrops, glacier, and a no information available class (NA-Area). We use multi-class segmentation labels for our experiments.

\subsection{Evaluation Metrics}
Gourmelon et al.~\cite{essd-14-4287-2022} offer unified postprocessing tools for multi-class segmentation outputs and evaluation metrics for research approaches within the CaFFe~\cite{essd-14-4287-2022} dataset. The precision, recall, F1-score, and intersection over union (IoU) are the metrics that reflect segmentation performance. In particular, the precision denotes the percentage of correctly predicted pixels among all positively predicted pixels while recall is the percentage of all positive pixels predicted to be positive. The F1 score is the harmonic mean of recall and precision. The IoU denotes the intersection between the pixels of the predicted class and the pixels of the actual target divided by their union. Formally, the four metrics are defined as follows:
\begin{equation}
\mathrm{Precision=\frac{TP}{TP+FP}}
\end{equation}
\begin{equation}
\mathrm{Recall=\frac{TP}{TP+FN}}
\end{equation}
\begin{equation}
\mathrm{F1\mbox{-}score=2\times\frac{Precision\times Recall}{Precision+Recall}}
\end{equation}
\begin{equation}
\mathrm{IoU=\frac{TP}{TP+FP+FN}}
\end{equation}
where $\mathrm{TP}$ indicates true positives. $\mathrm{FP}$ indicates false positives. $\mathrm{TN}$ indicates true negatives. $\mathrm{FN}$ indicates false negatives.

The mean distance error (MDE), is the most important evaluation metric for glacier front delineation. Postprocessing is needed for glacier front delineation after receiving multi-class segmentation predictions. Specifically, the Connected Component Analysis (CCA) is conducted on the merged multi-class segmentation prediction to receive all connected non-ocean regions and mark all but the largest connected component in the merged prediction leaving us with one ocean area. The boundary between this ocean area and all adjacent predicted glacier zones produces the 1-pixel-wide glacier termini. For further postprocessing details, we kindly refer the reader to~\cite{essd-14-4287-2022}. The MDE calculates the mean distance between the predicted and the ground truth calving front in meters. It is defined as follows:
\begin{multline}
    \mathrm{MDE}(I)=\frac{1}{\sum_{(P,Q)\in{I}}{(|P|+|Q|)}}\ast \\ \sum_{(P,Q)\in{I}}\bigl(\sum_{\bm{p}\in{P}}\min\limits_{\bm{q}\in{Q}}(||\bm{p}-\bm{q}||_2) + \\ \sum_{\bm{q}\in{Q}}\min\limits_{\bm{p}\in{P}}(||\bm{p}-\bm{q}||_2)\bigr)
\end{multline}
where $I$ is the set of all evaluated images, $P$ is the ground truth front pixels of one specific detected glacier front image, and $Q$ is the corresponding predicted front pixels to that image. $|*|$ is the cardinality of a set.

\subsection{Evaluation Protocol}
\subsubsection{Training}
The model weights are optimized by the AdamW~\cite{kingma2014adam} optimizer with an initial learning rate of 0.001. The learning rate exponentially decays with a parameter of 0.99. We train the model five times and for 300 epochs with a batch size of 30. The input image sizes for the target and context branches are both $288\times288$ pixels. We augment the input image online with multiple random rotations of 90$^\circ$ and horizontal/vertical flips each with a probability of 0.5. The weight parameters $\lambda{_1},\lambda{_2}, \lambda{_3}$ in~\cref{eq7} are set to 1.0, 1.0, and 0.5, respectively. The experiments are performed on a server equipped with an AMD EPYC 7662@2.0 GHz CPU and a single Nvidia A100-SXM4 GPU. The training process and front delineation details of our AMD-HookNet are shown in Algorithm~\ref{algo1}. For simplicity, we ignore the skip-connection representation in each U-Net architecture.

\begin{algorithm}[t]
\caption{Training process and front delineation details of AMD-HookNet}
\label{algo1}
\begin{algorithmic}[1]
\Require {$\mathbf{I}=\{\mathrm{T}_{i},\mathrm{C}_{i}~|~i\in R\}$~$\Leftarrow$ $i$ th pair of input images with sizes of $288\times288$ for target branch $\mathrm{T}$ and context branch $\mathrm{C}$, $R$ indicates the extracted patch number of one complete image $\mathbf{I}$ after using sliding window}
\Ensure {$\mathbf{M}$~$\Leftarrow$~Front delineation for $\mathbf{I}$}
\While{$i \leq R$}
\State{$\mathrm{C1}_{i}=$~Context-encoder$(\mathrm{C}_{i})$}
\State{$\mathrm{T1}_{i}=$~Target-encoder$(\mathrm{T}_{i})$}
\For{$D \in$~Convolutional blocks of decoder}
\State{$\mathrm{C2}_{i}^{D}=$~Context-decoder$^D(\mathrm{C1}_{i})$}
\If{$D == 1$}
\State{$\mathrm{T2}_{i}^{D}=$~Attention-hooking$^{D}_{up}(\mathrm{T1}_{i}, \mathrm{C2}_{i}^{D})$}
\ElsIf{$D == 2$ or $D == 3$}
\State{$\mathrm{T2}_{i}^{D}=$~Attention-hooking$^{D}_{up}(\mathrm{T2}_{i}^{D-1}, \mathrm{C2}_{i}^{D})$}
\Else
\State{$\mathrm{T2}_{i}^{D}=$~Target-decoder$^D(\mathrm{T2}_{i}^{D-1})$}
\EndIf
\EndFor
\State{Context~$\Leftrightarrow$~Supervision\{$\mathrm{C2}_{i}^{4}$\}}
\State{Target~$\Leftrightarrow$~Deep supervision\{$\mathrm{T2}_{i}^{1}, \mathrm{T2}_{i}^{2}, \mathrm{T2}_{i}^{3}, \mathrm{T2}_{i}^{4}$\}}
\State{$\mathrm{M}_{i}=\{\mathrm{T2}_{i}^{4}$\}}
\EndWhile
\State{$\mathbf{M}=$~CCA$\{\sum_{Merge}{\mathrm{M}_{i}}~|~i\in R\}$}
\end{algorithmic}
\end{algorithm}

\subsubsection{Testing}
To evaluate the generality and performance of our AMD-HookNet and to make fair performance comparisons, we use the test set and evaluation criteria in~\cite{essd-14-4287-2022} for testing, which includes the postprocessing for detection of glacier fronts and the detailed analysis of the statistical results. The performances of the best models within each training round are summarized and averaged to calculate the evaluation metrics.

\begin{table*}[t]
	\centering
	\caption{Comparisons between baseline and AMD-HookNet based on the evaluation metric mean distance error (MDE) in meters. Results are broken down by glacier and season. $\varnothing$ indicates the number of predictions that fail to detect a front. The number after $\in$ denotes the total number of images in the specific category (given glacier and season) in the test set.}
	\begin{tabular*}{\textwidth}{@{\extracolsep{\fill}}cccccccccc@{\extracolsep{\fill}}}
	\toprule
    \multirow{2}*{Glacier}&\multirow{2}*{Method}&\multirow{2}*{\bfseries MDE$\downarrow$}&\multirow{2}*{\bfseries $\varnothing$}&\multicolumn{2}{c}{Summer}&\multicolumn{2}{c}{Winter}\\
    & & & &\multicolumn{1}{c}{\bfseries MDE$\downarrow$}&\multicolumn{1}{c}{\bfseries$\varnothing$}&\multicolumn{1}{c}{\bfseries MDE$\downarrow$}&\multicolumn{1}{c}{\bfseries$\varnothing$} \\
	\midrule
	\multirow{2}*{All}&Baseline&753$\pm$76&1$\pm$1$\in$122&732$\pm$93&1$\pm$1$\in$68&776$\pm$65&0$\pm$0$\in$54 \\
	&Ours&\bfseries{438$\pm$22}&0$\pm$1$\in$122&\bfseries{374$\pm$39}&0$\pm$1$\in$68&\bfseries{495$\pm$44}&0$\pm$0$\in$54 \\
	\multirow{2}*{Columbia}&Baseline&840$\pm$84&0$\pm$0$\in$65&854$\pm$111&0$\pm$0$\in$28&826$\pm$66&0$\pm$0$\in$37 \\
	&Ours&\bfseries{489$\pm$31}&0$\pm$1$\in$65&\bfseries{435$\pm$48}&0$\pm$1$\in$28&\bfseries{533$\pm$49}&0$\pm$0$\in$37 \\
	\multirow{2}*{Mapple}&Baseline&287$\pm$48&0$\pm$1$\in$57&262$\pm$29&0$\pm$1$\in$40&340$\pm$93&0$\pm$0$\in$17 \\
	&Ours&\bfseries{164$\pm$28}&0$\pm$1$\in$57&\bfseries{174$\pm$32}&0$\pm$1$\in$40&\bfseries{152$\pm$45}&0$\pm$0$\in$17 \\
	\bottomrule
	\end{tabular*} \label{tab3}
\end{table*}
\subsection{Results}
The performance of the proposed method is compared with the state-of-the-art on the benchmark dataset CaFFe~\cite{essd-14-4287-2022}.

The segmentation performance and comparisons for precision, recall, F1-score, and IoU are presented in \cref{tab2}. Our AMD-HookNet is superior in all categories but the NA Area. Overall, it achieves a precision of 85.0$\pm$0.6, a recall of 85.0$\pm$0.7, an F1-score of 84.3$\pm$0.7 and an IoU of 74.4$\pm$1.0 on the entire test set, which outperforms the corresponding evaluation criteria of the baseline by \SI{0.8}{\percent}, \SI{5.4}{\percent}, \SI{4.2}{\percent} and \SI{4.7}{\percent}, respectively. AMD-HookNet significantly improves the performance of IoU score, which is considered the primary metric of segmentation tasks. We visualize the segmentation maps of the ground truth, the HookNet, and our AMD-HookNet in \cref{fig2}.

The glacier front delineation is performed after the postprocessing of the glacier segmentation predictions and the corresponding metric is the MDE. The comparisons of MDE broken down by glacier and season for baseline and AMD-HookNet are illustrated in \cref{tab3}. Note that the MDE on the complete test set of AMD-HookNet achieves 438$\pm$22\,m, obtaining an absolute performance gain of \SI{41.8}{\percent} over the baseline, indicating that the predicted glacier front of AMD-HookNet is closer to the ground truth. We visualize two accurate and two inaccurate predictions in \cref{fig3}. In this figure, the blue color represents ground truth, the yellow color indicates the detected glacier front, and the purple color shows the correct predictions, \ie the overlap between the prediction and the ground truth. In addition, AMD-HookNet obtains MDEs of 374$\pm$39\,m and 495$\pm$44\,m for summer and winter imagery, outperforming the baseline by \SI{51.2}{\percent} and \SI{43.2}{\percent}, respectively. This indicates that the model predicts more accurate on the summer imagery, when the ocean next to the glacier is not covered by sea ice. The ice melange which often covers the ocean in front of the terminus during winter has similar back scattering properties as the glacier, resulting in this performance drop between the seasons. With the performance gain of \SI{43.2}{\percent} over the baseline on wintertime images, AMD-HookNet also improves over the baseline on hard cases that feature ice melange.

\begin{table*}[t]
	\centering
	\caption{Comparisons between baseline and AMD-HookNet based on the evaluation metric mean distance error (MDE) in meters. Results are broken down by glacier and satellite. $\varnothing$ indicates the number of predictions that fail to detect a front. The number after $\in$ denotes the total number of images in the specific category (given glacier and satellite) in the test set.}
	\begin{tabular*}{\textwidth}{@{\extracolsep{\fill}}ccccccccc@{\extracolsep{\fill}}}
	\toprule
    Glacier&Method&&Sentinel-1&ENVISAT&ERS&PALSAR&TSX/TDX \\
	\midrule
	\multirow{4}*{All}&\multirow{2}*{Baseline}&{\bfseries MDE$\downarrow$}&2,201$\pm$246&493$\pm$119&404$\pm$172&437$\pm$42&547$\pm$61 \\
	&&{\bfseries$\varnothing$}&0$\pm$0$\in$33&0$\pm$0$\in$10&0$\pm$0$\in$2&0$\pm$0$\in$8&0$\pm$0$\in$69 \\
	&\multirow{2}*{Ours}&{\bfseries MDE$\downarrow$}&\bfseries{1,698$\pm$179}&\bfseries{386$\pm$126}&\bfseries{143$\pm$45}&\bfseries{322$\pm$202}&\bfseries{264$\pm$35} \\
	&&{\bfseries$\varnothing$}&0$\pm$1$\in$33&0$\pm$0$\in$10&0$\pm$0$\in$2&0$\pm$0$\in$8&0$\pm$0$\in$69 \\
	\multirow{4}*{Columbia}&\multirow{2}*{Baseline}&{\bfseries MDE$\downarrow$}&2,587$\pm$299&$\backslash$&$\backslash$&$\backslash$&587$\pm$67 \\
	&&{\bfseries$\varnothing$}&0$\pm$1$\in$18&$\backslash$&$\backslash$&$\backslash$&0$\pm$0$\in$47 \\
	&\multirow{2}*{Ours}&{\bfseries MDE$\downarrow$}&\bfseries{2,038$\pm$255}&$\backslash$&$\backslash$&$\backslash$&\bfseries{286$\pm$41} \\
	&&{\bfseries$\varnothing$}&0$\pm$0$\in$18&$\backslash$&$\backslash$&$\backslash$&0$\pm$0$\in$47 \\
	\multirow{4}*{Mapple}&\multirow{2}*{Baseline}&{\bfseries MDE$\downarrow$}&\bfseries{141$\pm$29}&493$\pm$119&404$\pm$172&437$\pm$42&246$\pm$57 \\
	&&{\bfseries$\varnothing$}&0$\pm$0$\in$15&0$\pm$0$\in$10&0$\pm$0$\in$2&0$\pm$0$\in$8&0$\pm$0$\in$22 \\
	&\multirow{2}*{Ours}&{\bfseries MDE$\downarrow$}&148$\pm$54&\bfseries{386$\pm$126}&\bfseries{143$\pm$45}&\bfseries{322$\pm$202}&\bfseries{116$\pm$25}  \\
	&&{\bfseries$\varnothing$}&0$\pm$1$\in$15&0$\pm$0$\in$10&0$\pm$0$\in$2&0$\pm$0$\in$8&0$\pm$0$\in$22 \\
	\bottomrule
	\end{tabular*} \label{tab4}
\end{table*}

The comparisons of the MDE broken down by glacier and satellite for baseline and AMD-HookNet are illustrated in \cref{tab4}. For Sentinel-1 satellite, AMD-HookNet achieves an MDE of 1698$\pm$179\,m, obtaining an absolute performance gain of \SI{22.9}{\percent} over the baseline. For ENVISAT satellite, AMD-HookNet achieves an MDE of 386$\pm$126\,m, obtaining an absolute performance gain of \SI{21.7}{\percent} over the baseline. For ERS satellite, AMD-HookNet achieves an MDE of 143$\pm$45\,m, obtaining a performance absolute gain of \SI{64.6}{\percent} over the baseline. For PALSAR satellite, AMD-HookNet achieves an MDE of 322$\pm$202\,m, obtaining an absolute performance gain of \SI{26.3}{\percent} over the baseline. For TSX/TDX satellite, AMD-HookNet achieves an MDE of 264$\pm$35\,m, obtaining an absolute performance gain of \SI{51.7}{\percent} over the baseline. Like with the baseline, the MDE is highest for Sentinel-1 images. We hypothesize that this originates from the glacier geometry more than from the sensor, as the high MDE is dominated by images of the Columbia Glacier where only one of the three calving fronts is identified, and the other two fronts negatively impact the MDE, as also explained in Gourmelon et al.~\cite{essd-14-4287-2022}.

\begin{table*}[t]
	\centering
	\caption{Comparisons between baseline and AMD-HookNet based on the evaluation metric mean distance error (MDE) in meters. Results are broken down by glacier and resolutio. $\varnothing$ indicates the number of predictions that fail to detect a front. The number after $\in$ denotes the total number of images in the specific category (given glacier and resolution) in the test set.}
	\begin{tabular*}{\textwidth}{@{\extracolsep{\fill}}cccccccccc@{\extracolsep{\fill}}}
	\toprule
    \multirow{2}*{Glacier}&\multirow{2}*{Method}&\multicolumn{2}{c}{20}&\multicolumn{2}{c}{17}&\multicolumn{2}{c}{7}\\
    & &\multicolumn{1}{c}{\bfseries MDE$\downarrow$}&\multicolumn{1}{c}{\bfseries$\varnothing$}&\multicolumn{1}{c}{\bfseries MDE$\downarrow$}&\multicolumn{1}{c}{\bfseries$\varnothing$}&\multicolumn{1}{c}{\bfseries MDE$\downarrow$}&\multicolumn{1}{c}{\bfseries$\varnothing$} \\
	\midrule
	\multirow{2}*{All}&Baseline&1,939$\pm$220&0$\pm$0$\in$45&437$\pm$42&0$\pm$0$\in$8&547$\pm$61&0$\pm$0$\in$69 \\
	&Ours&\bfseries{1,508$\pm$157}&0$\pm$1$\in$45&\bfseries{322$\pm$202}&0$\pm$0$\in$8&\bfseries{264$\pm$35}&0$\pm$0$\in$69 \\
	\multirow{2}*{Columbia}&Baseline&2,587$\pm$299&0$\pm$0$\in$18&$\backslash$&$\backslash$&587$\pm$67&0$\pm$0$\in$47 \\
	&Ours&\bfseries{2,038$\pm$255}&0$\pm$1$\in$18&$\backslash$&$\backslash$&\bfseries{286$\pm$41}&0$\pm$0$\in$47 \\
	\multirow{2}*{Mapple}&Baseline&323$\pm$69&0$\pm$0$\in$27&437$\pm$42&0$\pm$0$\in$8&246$\pm$57&0$\pm$0$\in$22 \\
	&Ours&\bfseries{248$\pm$64}&0$\pm$1$\in$27&\bfseries{322$\pm$202}&0$\pm$0$\in$8&\bfseries{116$\pm$25}&0$\pm$0$\in$22 \\
	\bottomrule
	\end{tabular*} \label{tab5}
\end{table*}

There are 3 different resolutions in the test set of CaFFe~\cite{essd-14-4287-2022}. 
The comparisons of the MDE broken down by glacier and resolution for baseline and AMD-HookNet are illustrated in \cref{tab5}. For a resolution of \SI{20}{m}, AMD-HookNet achieves an MDE of 1508$\pm$157\,m, obtaining an absolute performance gain of \SI{22.2}{\percent} over the baseline. For a resolution of \SI{17}{m}, AMD-HookNet achieves an MDE of 322$\pm$202\,m, obtaining an absolute performance gain of \SI{26.3}{\percent} over the baseline. For a resolution of \SI{7}{m}, AMD-HookNet achieves an MDE of 264$\pm$35\,m, obtaining an absolute performance gain of \SI{51.7}{\percent} over the baseline. The performance of AMD-HookNet with the spatial resolution of \SI{7}{m} is equipped with the lowest MDE.

\begin{table*}[t]
	\centering
	\caption{Comparisons on the ablation studies of AMD-HookNet.}
	\begin{tabular*}{\textwidth}{@{\extracolsep{\fill}}clcccccc@{\extracolsep{\fill}}}
	\toprule
    Glacier&Method&{\bfseries Precision$\uparrow$}&{\bfseries Recall$\uparrow$}&{\bfseries F1-score$\uparrow$}&{\bfseries IoU$\uparrow$}&{\bfseries{MDE} [m] {$\downarrow$}}\\
	\midrule
	\multirow{6}*{All}&Baseline&84.2$\pm$0.5&79.6$\pm$0.9&80.1$\pm$0.5&69.7$\pm$0.6&753$\pm$76 \\
	&HookNet&84.4$\pm$0.5&82.2$\pm$0.8&82.3$\pm$0.5&72.0$\pm$0.7&588$\pm$33 \\
	&HookNet$+$attention&84.4$\pm$0.8&82.6$\pm$1.0&82.4$\pm$0.6&72.1$\pm$0.5&500$\pm$80 \\
	&HookNet$+$deep supervision&84.5$\pm$0.5&83.6$\pm$0.5&83.1$\pm$0.2&72.9$\pm$0.1&510$\pm$\phantom{0}9 \\
	&HookNet$+$multi-hooking$+$deep supervision&84.8$\pm$0.5&83.9$\pm$0.7&83.5$\pm$0.3&73.4$\pm$0.2&489$\pm$79 \\
	&AMD-HookNet&\bfseries{85.0$\pm$0.6}&\bfseries{85.0$\pm$0.7}&\bfseries{84.3$\pm$0.7}&\bfseries{74.4$\pm$1.0}&\bfseries{438$\pm$22}\\
	\bottomrule
	\end{tabular*} \label{tab6}
\end{table*}

\begin{figure}[t]
\centering
\includegraphics[width=.39\textwidth]{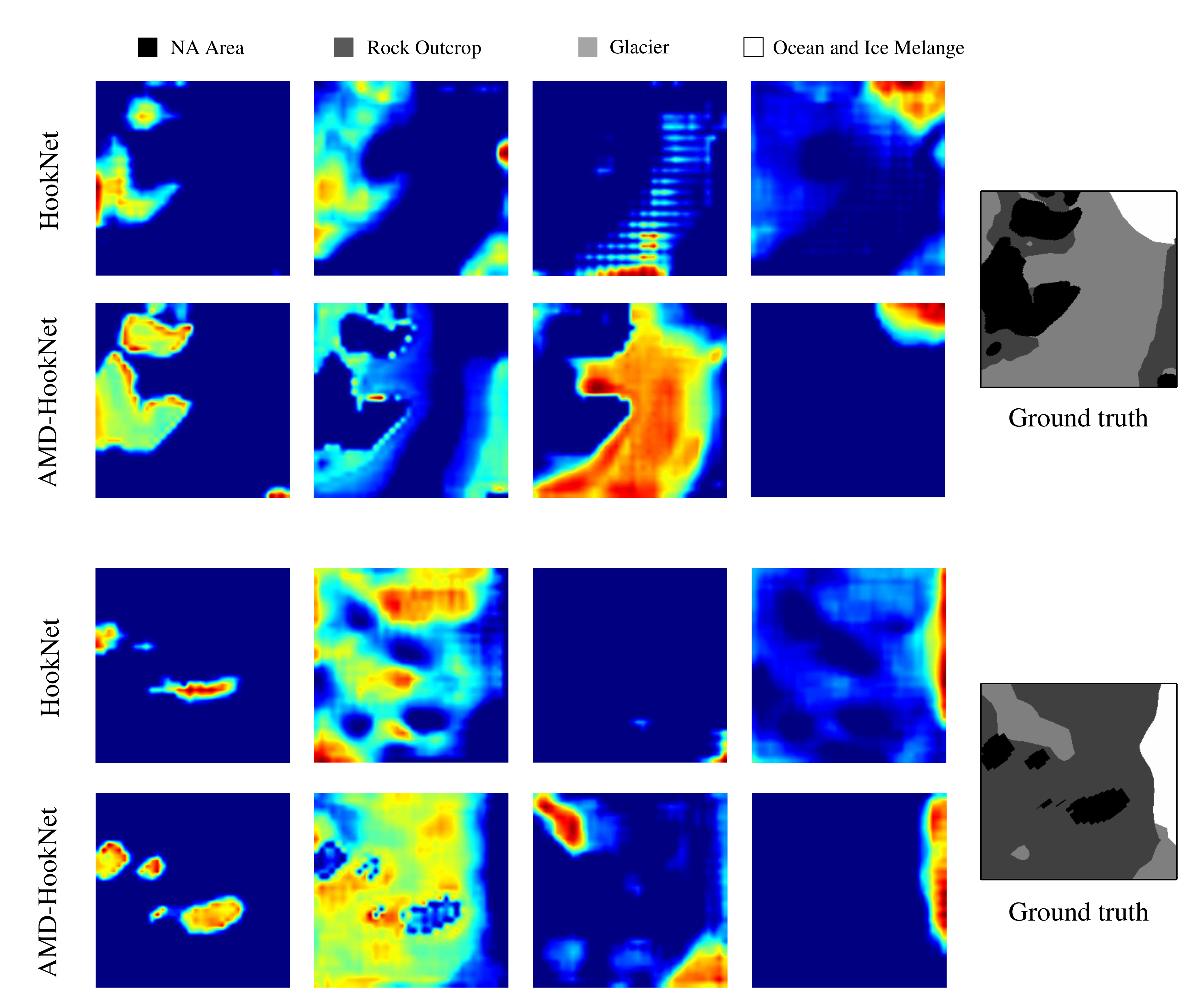}
\caption{Attention maps of the same hooked position (the first hooking) from HookNet and AMD-HookNet. Best viewed with zoom in.}\label{figh}
\end{figure}

\subsection{Ablation Study}
To fully investigate the effectiveness of individual components in the proposed approach, we conduct four ablation studies for the proposed components: (1)~base model: HookNet~\cite{van2021hooknet}, (2)~base model $+$ attention, the self-attention module is added only to one hooking operation in~(1), (3)~base model $+$ deep supervision, the deep supervision is added only to hooking operation in~(1), (4)~base model $+$ multi-hooking $+$ deep supervision, the multi-hooking operation and the corresponding deep supervision are added based on~(1), (5)~the proposed method: AMD-HookNet which incorporates the attention mechanism into~(4). 

The ablation studies for the components of AMD-HookNet are shown in \cref{tab6}. The highest influence comes from the use of the HookNet architecture itself reducing the MDE by \SI{22}{\percent} compared with the baseline. The use of attention seems more important than deep supervision. Combining all the components improves the results in comparison to the baseline by \SI{42}{\percent} and in comparison to the HookNet by \SI{26}{\percent}, respectively. This justifies the incorporation of all presented components into the final proposed AMD-HookNet.

In addition, we visualize the attention maps of the same hooked position (the first hooking) in HookNet and AMD-HookNet, as shown in \cref{figh}. The attention maps of multi-class segmentation are regarded as several one vs.\ others segmentations for visualization. It can be observed that AMD-HookNet has significantly improved feature aggregation capabilities compared with the original HookNet.

\section{Conclusion}\label{sec:conclusion}
In this paper, we propose a novel U-Net architecture for glacier calving front segmentation, which integrates attention mechanisms into the multi-hooking U-Net with deep supervision on the feature pyramid, dubbed as AMD-HookNet. It combines the global and local information of two individual U-Nets, and leverages the attention mechanism to make the network automatically focus on the valuable information in the combined feature maps improving the segmentation performance. Our experience from manual calving-front mapping supports these conclusions, since changing the zoom level helps to keep an overview of the general glacier geometry, when manually mapping challenging calving front sections. In addition, we introduce the multi-hooking operation with deep supervision to further optimize the training process. Extensive experimental performance comparisons demonstrate that our methodology outperforms the current state of the art by a large margin advancing the glacier segmentation task. More importantly, the success achieved in contextually federated network motivates us to investigate different information integration strategies and explore the potential of interactive attention mechanisms for vision transformer network in our future works.

\section*{Acknowledgments}
Grateful acknowledgment is made to the German Aerospace Center (DLR), the European Space Agency (ESA), and the Alaska Satellite Facility (ASF) for providing the SAR data for this study. The authors thank FAU Erlangen-Nürnberg and STAEDLER Foundation for financial support of this study under the Emerging Field Initiative TAPE: Tapping the Potential of Earth Observation. Moreover, the authors thank the Bavarian State Ministry of Science and the Arts for financial support within the International Doctorate Program “Measuring and Modelling Mountain glaciers and ice caps in a Changing ClimAte (M³OCCA)” by the Elite Network of Bavaria. The authors gratefully acknowledge the scientific support and HPC resources provided by the Erlangen National High Performance Computing Center (NHR@FAU) of the Friedrich-Alexander-Universität Erlangen-Nürnberg (FAU). The hardware is funded by the German Research Foundation (DFG).

\section*{IMPLEMENTATION}
Our codes that we used to produce the experimental results in this work will be available on GitHub~\cite{WinNT}.

\bibliographystyle{IEEEtran}
\bibliography{References}

\end{document}